\documentclass{article}

 \usepackage[preprint]{nips_2018}

\usepackage{amsmath, amsthm, amssymb, graphicx, verbatim} 
\usepackage{bbm}
\usepackage[utf8]{inputenc} 
\usepackage[T1]{fontenc}    
\usepackage{hyperref}       
\usepackage{url}            
\usepackage{booktabs}       
\usepackage{amsfonts}       
\usepackage{nicefrac}       
\usepackage{microtype}      

\newtheorem{theorem}{Theorem}
\newtheorem{proposition}[theorem]{Proposition}

\theoremstyle{definition}
\newtheorem{definition}[theorem]{Definition}

\theoremstyle{remark}

\title{On Mechanisms for Transfer using Landmark Value Functions in Multi-Task Lifelong Reinforcement Learning}

\author{
  Nick Denis
  \\
  Department of Mathematics and Statistics\\
  University of Ottawa\\
  Ottawa, ON K1N 6N5\\
  \texttt{ndeni032@uottawa.ca}
}

\begin{document}

\maketitle

\begin{abstract}
Transfer learning across different reinforcement learning (RL) tasks is becoming an increasingly valuable area of research. We consider a goal-based multi-task RL framework and mechanisms by which previously solved tasks can reduce sample complexity and regret when the agent is faced with a new task. Specifically, we introduce two metrics on the state space that encode notions of traversibility of the state space for an agent. Using these metrics a topological covering is constructed by way of a set of \textit{landmark} states in a fully self-supervised manner. We show that these landmark coverings confer theoretical advantages for transfer learning within the goal-based multi-task RL setting. Specifically, we demonstrate three mechanisms by which landmark coverings can be used for successful transfer learning. First, we extend the Landmark Options Via Reflection (LOVR) framework to this new topological covering; second, we use the landmark-centric value functions themselves as features and define a greedy \textit{zombie policy} that achieves near oracle performance on a sequence of zero-shot transfer tasks; finally, motivated by the second transfer mechanism, we introduce a learned reward function that provides a more dense reward signal for goal-based RL. Our novel topological landmark covering confers beneficial theoretical results, bounding the $Q$ values at each state-action pair. In doing so, we introduce a mechanism that performs action-pruning at infeasible actions which cannot possibly be part of an optimal policy for the current goal. In effect, this action-pruning mechanism can be interpreted as a sense of \textit{safety}, as it prevents the agent from taking any actions that are sufficiently detrimental towards accomplishing its goal. Empirical results on the cliff-walker domain support these theoretical results. Finally, since each of the transfer mechanisms each have their own benefits and deficits in terms of finite-time analysis, we demonstrate empirically that a hierarchical agent that uses a multi-armed bandit controller at the start of each episode to select from these transfer mechanisms learns to use the transfer mechanisms when they confer the most advantage, and discontinue their use when no longer advantageous.

\end{abstract}


\section{Introduction}
\vspace{-0.5em}
The strength of reinforcement learning (RL) in solving sequential decision problems is apparent in the broad and increasing range of problem-types considered by RL researchers. One example of these are lifelong learning settings [1]. They assume an RL agent is faced with a sequence of MDPs that share certain properties and are thus viewed as coming from a common {\it environment}. As in transfer learning[2],
rather than beginning fresh with each task, an efficient lifelong agent should be able to leverage past knowledge of its experience within the environment in order to solve each new task as quickly as possible.
Hierarchical RL and the use of temporally extended actions are essential in overcoming the curse of dimensionality in many settings [3]. How to structure control and learning within an RL hierarchy are active current research questions [4-7].
More generally - hierarchical methods, representation learning and semi-supervised learning (SSL) are all well-established areas of research in their own right (see resp. [8,9] for surveys), with fruitful interplay between them, e.g., [10,11] for two recent examples. 

We address how previously solved tasks can be used to reduce sample complexity and/or regret on future tasks sampled from the same environment. Specifically, we ask: how can access to optimal Q value functions with respect to a set of landmark states help in solving a new goal-based task from the same environment? To this end, we extend the Landmark Options Via Reflection (LOVR) framework by replacing the notion of an $\eta$-reachability covering of landmarks, $\mathbb{L}$, with a true topological covering under metrics on $\mathcal{S}$ that carry meaningful semantics for the goal-based multi-task setting. This landmark covering is first learned in a purely self-supervised setting, after which we introduce different mechanisms by which $\{Q^{\ell}\}_{\ell\in \mathbb{L}}$ can be used to solve a new task. The first mechanism simply extends the LOVR framework [12]. The second mechanism involves representing the current state $s \in \mathcal{S}$ as $\{V^{\ell}(s)\}_{\ell \in \mathbb{L}} \equiv V^{\mathbb{L}}(s) \equiv (V^{\ell_{1}}(s), V^{\ell_{2}}(s), ..., V^{\ell_{\vert \mathbb{L} \vert}}(s))$. This approach finds inspiration from the philosophy of phenomenology, specifically that of Martin Heidegger, whereby our representation of objects in the world are not \textit{objective} in any sense, but rather come to us by how the object provides us subjective utility, being \textit{ready-to-hand}. Hence, this state-representation, $V^{\mathbb{L}}(s)$, represents the current state of the environment in terms of value or utility from a vector of contexts. Depending on the environment, if $\mathcal{S} = \mathbb{R}^{d}$, and $\vert \mathbb{L} \vert < d$, then $V^{\mathbb{L}}(s)$ has the benefit of inducing a dimensionality reduction on the state representation. In this study, this state representation is coupled to a greedy and deterministically defined \textit{zombie} policy, which astonishingly empirically achieves near-oracle levels of regret on future tasks \textit{despite no further learning updates taking place}. This represents a truly effective zero-shot transfer learning approach. Finally, the third mechanism for transfer introduced is motivated by the $V^{\mathbb{L}}$ state representation to define a novel reward function which is more dense and varied than the action-penalty reward structure normally considered. We also show that a hierarchical agent comprising of a bandit-controller which at the start of each episode selects amongst baseline learning and the transfer mechanisms achieves state of the art performance.

From a theoretical standpoint under the topological landmark covering we show that for any new goal task $g$, lower and upper bounds on the optimal state-action value function can be made. In doing so, action-pruning can be used to prevent actions at particular states that cannot possibly realize the optimal state-action value function for the given task. In effect, this action-pruning mechanism can be interpreted as a sense of \textit{safety}, as it prevents the agent from taking any actions that may be detrimental towards accomplishing its goal. Empirical results on the cliff-walker domain support these theoretical results.

Our contributions include 
\begin{itemize}
\item We introduce two metrics on $\mathcal{S}$ that are semantically meaningful for the goal-based multi-task RL setting.
\item We extend the LOVR framework using a landmark covering, $\mathbb{L}$, which is a true topological covering of $\mathcal{S}$.
\item For any new goal task, $g$, we show that $\mathbb{L}$ can be used to obtain theoretical bounds on $V^{g^{*}}(s)$, $\forall s \in \mathcal{S}$. 
\item The aforementioned theoretical bounds are used to motivate an \textit{action-constrained} exploration strategy whereby actions that cannot possibly be taken by an optimal policy are not considered during exploration. This action-constrained exploration approach can be interpreted as a sense of \textit{safety} in exploration, which is demonstrated empirically in the cliff-hanger domain.
\item We introduce a novel state representation approach by way of $V^{\mathbb{L}} \equiv \{V^{\ell}(s)\}_{\ell \in \mathbb{L}}$.
\item A greedy and deterministic \textit{zombie} policy is introduced, which, when coupled with the the $V^{\mathbb{L}}$ state representation, achieves near oracle regret on future tasks, despite no further learning updates being made.
\item A novel form of transfer is introduced through a reward function that utilizes $V^{\mathbb{L}}$, and empirically achieves lower finite-time regret compared to baseline.
\item A hierarchical agent utilizing a bandit controller to select among the transfer mechanisms and the baseline learning algorithm makes use of each approach at distinct periods of the learning process leading to strong empirical performance. 
\end{itemize}

For the remainder of the paper we review relevant background and related research, including the LOVR framework. Next, we introduce the metrics on $\mathcal{S}$ used in this study, discuss the topological covering $\mathbb{L}$ and the theoretical results stemming from its use in the goal-based multi-task RL setting, including the novel action-constrained exploration strategy. We demonstrate empirical results for the action-constrained exploration strategy on the cliff-walk domain, before introducing the $V^{\mathbb{L}}$ state representation, \textit{zombie} policy, $\pi^{\mathbb{L}}$ and transfer through reward function $r^{\mathbb{L}}$. Empirical results on the MNIST world domain are stated.

\section{Background}\label{sec:background}
\vspace{-0.5em}
\paragraph{Goal-based multi-task reinforcement learning}

We consider a multi-task RL setting where the agent is confined to a stationary environment, and throughout its lifetime is assigned a sequence of tasks. A finite sequence of tasks are represented as  episodic MDPs $\mathcal{M}_{i} = \langle \mathcal{S}, \mathcal{A}, P, \mathcal{R}, s_{g_{i}}, \nu \rangle$, $i \in [T]$, with respective terminal states $s_{g_i}$. In summary: the state space $\mathcal{S}$, set of actions $\mathcal{A}$, transition probability kernel $P$, reward function $\mathcal{R}$ and initial state distribution $\nu$ all remain fixed across tasks and only the goal state varies. The goal state $s_{g_{i}}$ thus encodes the $i$'th task, though when speaking of a single task denoted as simply $s_g$. In many settings it may be more applicable to consider goal sets $\mathcal{G}_{i} \subset \mathcal{S}$, rather than a single goal state. Our usage of terminal state follows the common formulation [13]for episodic tasks where one wants the agent, such as a robot, to perform a single specific function, and upon completion to become inactive and remain so until a new episode or task is begun.
In general one could let $K_{i}$ be the number of episodes that $\mathcal{M}_{i}$ is run before the next task is assigned, but we will take $K_i \equiv K$ to be independent of $i$ in the present paper.  We consider the action-penalty reward structure [14] of $-1$ for all transitions, which reinforces the agent towards a policy of arriving at the current goal in as few steps as possible. The goal of the lifelong learning agent is to solve for a sequence of policies $\{\pi^{*}_{i}\}_{i=1}^{T}$ which minimize cumulative regret. It is clear that under this reward structure that $V^{g}(s)$ is the negative expected number of steps from state $s$ to goal $g$, and since $\pi^{*}$ is optimal, $V^{g^{*}}(s)$ encodes the ``fastest" path to the goal.

\paragraph{Options Framework}
The options framework is a mathematically principled approach for temporally extended actions [15]. An option $o \in \mathcal{O}$ is a triple, $o = \langle I_{o}, \pi_{o}, \beta_{o} \rangle$, where $\mathcal{I}_{o} \subseteq \mathcal{S}$ is the initiation set, $\pi_{o}$ is the option policy that maps states to primitive actions, and $\beta_{o}$ is the termination function  which controls when to terminate $\pi_{o}$ and return control back to $\pi$. The inclusion of options in a MDP results in a Semi-MDP (SMDP), where standard RL algorithms apply in the SMDP setting [15]. Landmark options were first introduced in [16,15] as policies leading towards  ``landmark" states. We retain this aspect but the way they are employed in our framework is different [12].

\section{Introduced Traversibility Metrics and Landmark Coverings}
\vspace{-0.5em}

\paragraph{}We begin by introducing two metrics on the state space of MDPs. These metrics consider first hitting times between states, which carry with them a meaningful and useful interpretation under goal based RL. Before introducing the two metrics, we recall the notion of a topological covering.

\begin{definition}(Covering) Let $(\mathcal{X}, d)$ be a metric space. Let $\eta > 0$. We say $\mathcal{C} \subset \mathcal{X}$ is an $\eta$-cover of $\mathcal{X}$ if $\forall x \in \mathcal{X}$, $\exists c \in \mathcal{C}$ such that $d(x,c) \leq \eta$.
\end{definition}

\paragraph{}Recall that for all tasks, $g$, we use the action-penalty reward function $r \equiv -1$. It is easy to show that under this reward function, $\forall g, s \in \mathcal{S}$, $V^{g^{*}}(s) = - \mathbb{E}_{\pi^{*}}[\text{min. number of steps from s to g}]$. That is, $\vert V^{g^{*}}(s)\vert$ is the expected first hitting time of state $g$ from state $s$ under some optimal shortest path policy. With this, we introduce two metrics.

\begin{definition}(Round trip metric) For $d_{rt}: \mathcal{S} \times \mathcal{S} \to [0,2D]$, $D$ the diameter of $\mathcal{S}$, we define the round trip function $d_{rt}$:
\begin{align*}
d_{rt}(x,y) = \vert V^{x^{*}}(y) \vert + \vert V^{y^{*}}(x) \vert
\end{align*}
\end{definition}

\begin{definition}(Max one-way metric) For $d_{\infty}: \mathcal{S} \times \mathcal{S} \to [0,D]$, $D$ the diameter of $\mathcal{S}$, we define the max one-way function $d_{\infty}$:
\begin{align*}
d_{\infty}(x,y) = max\{\vert V^{x^{*}}(y) \vert ,\vert V^{y^{*}}(x) \vert\}
\end{align*}
\end{definition}

\paragraph{}It is straight forward to prove that both the round-trip and max one-way functions are in fact metrics on $\mathcal{S}$. These metrics are helpful for goal-based RL as they carry meaningful semantics no matter the state representation, as opposed to other metrics such as $\mathcal{L}_{p}$ norms, which don't carry any meaning in state representation spaces (e.g. images). Using these metrics we look to construct $\eta$-coverings of $\mathcal{S}$ and see how they can confer theoretical benefits in the goal-based multi-task RL setting. First, we show that an agent with access to such a covering, $\mathbb{L}$, one is able to immediately bound the value of any state $s \in \mathcal{S}$ for any novel task $g$.

\vspace{0.2cm}

\begin{proposition}Let $\mathbb{L}$ be an $\eta$-cover of $\mathcal{S}$ under the round-trip metric $d_{rt}$. Let $g \in \mathcal{S}$ be a task. Then $\exists \ell \in \mathbb{L}$ such that $\vert V^{g^{*}}(\ell) \vert \leq \eta - \vert V^{l^{*}}(g) \vert$, and $\forall s \in \mathcal{S}$ $\exists \ell \in \mathbb{L},$:
\begin{align*}
\vert V^{g^{*}}(s) \vert \in \Big[\text{max}\{0, \vert V^{\ell ^{*}}(s) \vert - \vert V^{\ell^{*}}(g) \vert\},   \vert V^{\ell ^{*}}(s) \vert - \vert V^{\ell^{*}}(g) \vert\ + \eta\Big]
\end{align*}
Moreover these bounds are tight, in the sense that $\exists$ an MDP  $\mathcal{M}$ such that these bounds are realized.
\end{proposition}

\vspace{0.2cm}

\begin{proposition}Let $\mathbb{L}$ be an $\eta$-cover of $\mathcal{S}$ under the max one-way metric $d_{\infty}$. Let $g \in \mathcal{S}$ be a task. Then $\exists \ell \in \mathbb{L}$ such that $\vert V^{g^{*}}(\ell) \vert \leq \eta$, and $\forall s \in \mathcal{S}$ $\exists \ell \in \mathbb{L},$:
\begin{align*}
\vert V^{g^{*}}(s) \vert \in \Big[\text{max}\{0, \vert V^{\ell ^{*}}(s) \vert - \vert V^{\ell^{*}}(g) \vert\},  \vert V^{\ell ^{*}}(s) \vert\ + \eta  \Big]
\end{align*}
Moreover these bounds are tight, in the sense that $\exists$ an MDP $\mathcal{M}$ such that these bounds are realized.
\end{proposition}

\paragraph{}Note that for both Proposition 4 and 5, the $\ell \in \mathbb{L}$ referenced in the existence statements is the landmark that \textit{witnesses} the goal, that is $d(g,\ell) \leq \eta$.

\section{Proposed Transfer Mechanisms}\label{sec:framework}
\vspace{-0.5em}

\subsection{Transfer through Safety via Action Pruning}

\paragraph{}The first transfer mechanism we introduce relates to a notion of safety during exploration. Having bounds on the optimal value  function, and thus the expected first hitting time along some optimal policy, of any novel goal task $g$ is quite beneficial. By having both upper and lower bounds on the expected first hitting time for some optimal policy, inductive biases can be encoded into exploration policies that can remove from consideration actions that cannot possibly be selected by an optimal policy. We believe that this encoding can be interpreted as a sense of \textit{safety} during exploration. Our aim is to introduce an action-pruning mechanism based on these theoretical bounds. We begin by motivating the intuition behind this action pruning mechanism.

\vspace{0.2cm}
\begin{definition}(Oracle function) Let $o:\mathcal{S}\times\mathcal{S} \to [0,D]$ be the oracle function, where $o(s, g) \equiv o_{g}(s) \equiv \vert V^{g^{*}}(s) \vert.$
\end{definition}

\paragraph{}Hence, the oracle function $o$ takes as arguments an arbtriary state-goal pair, and returns the expected first hitting time from the state to the goal under some optimal policy.

\vspace{0.2cm}

\begin{definition}(Oracle feasible actions) The oracle feasible actions, $A_{o}(s, g, \mathbb{L})$, are those actions that are feasible or plausibly possible with the optimal landmark $Q$ values and the oracle. Symbolically,
\begin{align*}
A_{o}(s, g, \mathbb{L}) := \{ a \in \mathcal{A} : \vert Q^{\ell^{*}}(s,a) \vert \leq \vert V^{\ell^{*}}(g) \vert + o(s), \ \forall \ell \in \mathbb{L} \}
\end{align*}
\end{definition}

\paragraph{}$A_{o}(s,g,\mathbb{L})$ containts all the feasible actions in the sense that if for some given goal $g$, and state $s$ if $\pi^{*}_{g}(s) = a$ then $a \in A_{o}(s,g, \mathbb{L})$, and equivalently if $a \notin A_{o}(s,g,\mathbb{L})$ then $\pi^{*}_{g}(s) \neq a$. The first claim is trivial to show since $\pi^{*}_{g}(s)$ is the action that maximizes $Q^{g^{*}}(s,a)$ which also realizes $o(s)$. To show that if a given action $a$, $a \notin A_{o}(s,g,\mathbb{L})$ then $\pi^{*}_{g}(s) \neq a$, suppose otherwise. Since $a \notin A_{o}(s,g,\mathbb{L})$ then $\exists \ell \in \mathbb{L}$ such that $\vert Q^{\ell^{*}}(s,a)  \vert > \vert V^{\ell^{*}}(g) \vert + \vert V^{g^{*}}(s) \vert$. However,  this implies that the expected number of steps to arrive at $\ell$ from $s$ by taking action $a$ is strictly greater than taking the optimal policy to $g$ then taking the optimal policy from $g$ to $\ell$. But this is a contradiction since $a$ is the optimal action under $\pi^{*}_{g}(s)$. We make no assumptions to having access to an oracle, $o$, therefore we do not have access to $A_{o}(s,g, \mathbb{L})$. However, we will construct sets of feasible actions based on each metric, which are supersets of $A_{o}$, and hence again any $a$ not in the feasible sets then cannot be part of any optimal policy. We do so by using our upperbound estimates on $o(s)$ as given by Propositions 4 and 5.

\vspace{0.2cm}

\begin{definition}(Feasible sets) Given $\mathbb{L}$ an $\eta$-cover of $\mathcal{S}$ under $d_{rt}$ or $d_{\infty}$, the feasible sets for each covering are defined, respectively:
\begin{align*}
A_{rt} &= \{a \in \mathcal{A} : \vert Q^{\ell^{*}}(s,a) \vert \leq  \vert V^{\ell ^{*}}(s) \vert + \eta, \ \forall \ell \in \{\ell' \in \mathbb{L} \vert \ d_{rt}(g,\ell') \leq \eta \} \}\\
A_{\infty} &= \{a \in \mathcal{A} : \vert Q^{\ell^{*}}(s,a) \vert \leq \vert V^{\ell^{*}}(g) \vert + \vert V^{\ell}(s) \vert + \eta, \ \forall \ell \in \{\ell' \in \mathbb{L} \vert \ d_{\infty}(g,\ell') \leq \eta \} \}\\
\end{align*}

\end{definition}

\paragraph{}We see that the covering $\mathbb{L}$ can now be used to make exploration more efficient by restricting action selection to the feasible actions. In many states it is likely that the feasible actions are simply the entire set of actions $\mathcal{A}$. However, for those actions that are not amongst the feasible actions, then these actions will, in expectation, delay the agent from arriving at the goal state significantly and hence should not be taken. Note that the feasible sets are defined purely with respect to (state) value functions associated to the landmark states $\mathbb{L}$, which can be solved in a self-supervised learning manner in an initial phase, before the agent is assigned the sequence of tasks. We first demonstrate how restricting exploration to only those actions within the feasible sets can confer notions of safety and thereby reduce sample complexity by performing experiments on the cliff walker domain (Results Section).

\

\subsection{Transfer through Landmark Options}

\paragraph{}The second transfer mechanism considered is simply an extension to the previously introduced Landmark Options Via Reflection (LOVR) framework [12]. A more detailed explanation can be found in [12]. Briefly, by properly defining the initiation set and termination conditions of the set of options, each landmark state is associated to a landmark-centric value function, $\{Q^{\ell}\}_{\ell \in \mathbb{L}}$, which induces a landmark option policy $o_{\ell}$. Specifically, at the start of each episode, the landmark option associated to the landmark state that is closest to the current goal $g$ under the $\eta$-covering is selected. The option is terminated when the current state is within an $\eta$-ball of the landmark closest to the current goal. Due to the metrics considered in these coverings, this also implies that the current goal $g$ is also within an $\eta$-ball around this landmark. Upon terminating the landmark option, the initiation set of all landmark options are all empty, and hence the agent can only select primitive actions. In essence, this encoding of LOVR drives the agent to an $\eta$-ball around the landmark state that is closest to the current goal, and while within this $\eta$-ball, explores and solves for an optimal policy towards the goal from this $\eta$-ball. Whenever the agent leaves the $\eta$-ball, it selects the landmark option that drives the agent back to the ball. In effect, LOVR reduces the state-space of the current task to an $\eta$-ball, where $\eta$ is a hyperparameter set by the experimenter. Rather than exploring over the entire state space $\mathcal{S}$, the agent performs exploration only within an $eta$ ball, which can be a dramatic reduction in size. By controlling the size of $eta$, the experimenter trades off an initial ``overhead" for learning optimal policies for the covering $\mathbb{L}$ with a reduced effective state space size on all future tasks. Prior work introducing LOVR showed a dramatic improvement in finite sample complexity over baseline methods for a lifelong multi-task RL agent [12].

\subsection{Transfer Through Intermediate State Representation: $V^{\mathbb{L}}$}

\paragraph{}Due to the semantic nature of the value functions considered in this setting, the evaluation of a single state with respect to the value functions of multiple previously solved tasks can be used to represent a state as a finite length vector of expected first arrival times to various landmark states that cover the environment. This state representation was inspired by the works of Martin Heidegger's \textit{Being  and Time}. All to briefly, Heidegger introduces the notion of objects being \textit{ready-to-hand}. Our experience of an object that is ready-to-hand is rather that we do not experience the object at all. We make no notice of a pen when we use it to write, rather it is simply an extension of ourselves. It is only when there is a malfunction (e.g. pen out of ink) that we become aware of the pen, as it has lost its ``pen-ness", by way of a loss of functionality. Objects that are ready-to-hand are extensions of ourselves, and we come to experience and view the world through the utility of these objects, just as we experience and view the world through the utility of our own bodies. In this way, we make use of Heidegger's phenomenology by considering a set of previously learned options as extensions of self. We encode this by representing the state not as it is given to the agent, but rather, as a vectorized representation of the values of the current state with respect to the previously learned options. In this way, the options can be interpreted as acting as an intermediate representation. This representation asks the question ``What value is this current state with respect to all of my skills and knowledge of the world?", and the answer being $V^{\mathbb{L}}(s)$.

\paragraph{}Given $\mathbb{L}$, we define $V^{\mathbb{L}}(s) = (V^{\ell_{1}}(s), V^{\ell_{2}}(s), ..., V^{\ell_{\vert \mathbb{L}\vert}}(s))$, for some fixed indexing of  $\mathbb{L}$. Hence, $V^{\mathbb{L}}$ is an $|\mathbb{L}|$ dimensional state representation that encodes semantics about the traversibility of the state space. $Q^{\mathbb{L}}$ is defined similarly. If $|\mathbb{L}|$ = 1, then $V^{\mathbb{L}}(s)$ representations may actually be harmful, as they may alias the state space since many different states can have (approximately) the same expected first hitting time to $\ell$, thus converting the problem to a partially observable MDP.  However, if $\mathbb{L}$ is {\it large} enough, it is fair to assume that the map $\phi(s) = V^{\mathbb{L}}(s)$ becomes injective, and hence the aliasing affect will no longer be an issue. Moreover, in settings where the goal state $g$ is provided to the agent, then the agent can represent this state as $V^{\mathbb{L}}(g)$. Such a representation can be favorable, as it can result in a dramatic dimensionality reduction of the state representation, especially in visual (image) environments. This can allow for smaller function approximation architectures to be used to speed up learning and also provide stability. Moreover, this state representation carries with it a natural and useful semantics, that of a localization of $s$ with respect to reference states $\ell_{1}, ..., \ell_{\mathbb{L}}$ in terms of expected first hitting times. We expect this state representation to be easy to exploit, both immediately by following deterministic greedy policies that minimize the distance between $s$ and $g$ in this representation space, and by use of function approximations such as neural networks which we believe may more quickly learn useful feature maps from this already semantically rich state representation. 

\subsubsection{Zombie Policy}

\paragraph{}
Using the $V^{\mathbb{L}}$ state representation, we consider a greedy heuristic policy termed the {\it zombie} policy. As discussed previously, if this $V^{\mathbb{L}}$ is an injective map on $\mathcal{S}$, then state aliasing is of no concern, and this state representation will still follow the Markov property. Moreover, this state representation carries with it semantic content relevant to the goal-based RL setting studied here. Noting that,
\begin{align*}
V^{\mathbb{L}}(s) - V^{\mathbb{L}}(g) = (0,0,...,0)  = \textbf{0} \iff V^{\mathbb{L}}(s) = V^{\mathbb{L}}(g) \iff s = g,
\end{align*}

we reason that a potentially simple and useful heuristic is to take actions that move the agent greedily towards $V^{\mathbb{L}}(g)$ in this representation space. We define the {\it zombie} policy, $\pi^{\mathbb{L}}_{\mathcal{Z}}$,
\begin{align*}
\pi^{\mathbb{L}}_{\mathcal{Z}}(s)  := \underset{a \in \mathcal{A}} {\mathrm{argmin}} \ \ || Q^{\mathbb{L}}(s,a) - V^{\mathbb{L}}(g)||_{1}.
\end{align*}

\paragraph{}This policy selects the action that is expcted to bring the agent towards a state that has similar expected first hitting times to each of the landmarks as the current goal does. Proper theoretical analysis is required, however our initial intuition is that with a sufficiently {\textit spread out} set of landmarks over $\mathcal{S}$, environments with low connectivity (e.g. each state has a small number of possible successor states) and injectivity of this representation, we believe this representation should confer strong benefits in terms of sample complexity and regret bounds. 

\paragraph{}We highlight that following this zombie policy no longer utilizes any learning updates, is completely deterministic, and hence is a purely \textit{zero-shot} transfer mechanism, as no further learning occurs on the target task.

\subsection{Transfer Through Reward Function}

\paragraph{}The final mechanism for transfer in goal-based multi-task RL considered here involves the agent implementing a highly dense reward function used to solve each subsequent task. This highly dense reward function is built from $V^{\mathbb{L}}$, and follows a similar line of reasoning as that used for the zombie policy $\pi^{\mathbb{L}}_{\mathcal{Z}}$. Often in RL environments such as robotics tasks, the reward function is defined as the (scaled) negative distance between the state and the target goal state. Though this reward function is highly dense, especially as compared to the action penalty reward function, in real world settings oracle knowledge of distances is not provided. Moreover, metrics on state spaces tend not to produce any meaningful semantics, as distance between two states in pixel space, for example, do not carry any semantics. However, under the framework presented here, the agent has learned a useful state representation, and can be used to define the distance $d^{\mathbb{L}}(s,g) := ||V^{\mathbb{L}}(s) - V^{\mathbb{L}}(g)||_{1}$, which measures, relative to the goal state $g$, what is the total {\it difference} in expected steps $s$ is to each $\ell \in \mathbb{L}$ relative to $g$. Hence, we use the transfer reward function $r^{\mathbb{L}}(s,a,s') = -\beta d^{\mathbb{L}}(s',g)$ in our experiments, for some $\beta > 0$. We note: $d^{\mathbb{L}}(s,g) = 0 \iff s= g$; unlike $\pi^{\mathbb{L}}_{\mathcal{Z}}$ which assumes injectivity of $V^{\mathbb{L}}$, and is not capable of any further learning, $r^{\mathbb{L}}(s,a,s')$ does not require injectivity of $V^{\mathbb{L}}$, rather, requires only the much weaker condition that $V^{\mathbb{L}}(s) = V^{\mathbb{L}}(g) \iff s=g$. Under this assumption, $r^{\mathbb{L}}(s,a,s')$ provides a dense reward at every step, and in minimizing the undiscounted sum of rewards the agent is not likely to solve for a policy that receives small, but negative rewards indefinitely, but will favor arriving at the goal state $g$, and do so ``quickly". For this reason, we believe that $r^{\mathbb{L}}(s,a,s')$ should work quite well, especially with the $V^{\mathbb{L}}$ state representation, over baseline methods and should asymptotically outperform $\pi^{\mathbb{L}}_{\mathcal{Z}}$. For our experiments we used $\beta = 0.1$.

\section{Results}\label{sec:results}
\vspace{-0.5em}

\subsection{Safety through Action-Pruning}

\paragraph{}To demonstrate how the theoretical results above can be translated into notions of safety, we use the feasible action set as defined above in the cliff-walker domain. Cliff-walker is a stochastic gridworld domain that simulates walking along a windy cliff. As seen in Figure 1, we use a $4 \times 12$ gridworld, where the cliff runs along the bottom row between the first and last columns. The agent is equipped with four actions, each associated to movement in a cardinal direction. The agent always begins each episode at the bottom left state. We simulate a \textit{windy} cliff-walker environment, where for all non-cliff states, the agent moves successfully according to the action selected with probability 0.8, but with probability 0.2 moves down, if possible. Moreover, if the agent selects the \textit{down} action, with equal probability the agent moves down one or two spaces (when possible). In this setup, the cliff acts as a \textit{sticky} state, where with probability 0.95, no matter what the action, the agent remains at that state, and with proability 0.05 the action taken is successful. We use the $d_{\infty}$ metric to cover $\mathcal{S}$ with $\eta = 8$. For this environment, this results in $\mathbb{L}$ comprising of three non-cliff states, and all the cliff-states (see Figure 1). The initial state is always the bottom left state, and we consider experiments where the goal task $g$ is arriving at the bottom right state. The optimal policy involves first moving upward, then moving as far right as possible. Moving downwards in any of the cliff columns is considered dangerous, especially in the bottom three rows. We run experiments with a baseline Q-learning agent (no transfer baseline), a Q-learning agent with action-pruning, a Q-learning LOVR agent and a Q-learning LOVR agent with action-pruning. Exploration is performed using $\epsilon$-greedy exploration with $\epsilon = 0.1$. Each implementation is run for $n=100$ experiments, each experiment lasting 1000 episodes, each episode lasting a maximum of 1000 time steps (or until the agent reaches the goal state). For each experiment we count the number of time steps the agent is on the cliff and use this as a measure for how safe the agent is during exploration. 

\paragraph{}The baseline Q-learning agent is on the cliff an average of 22754.31 time steps per experiment. For all other agents we report the relative percent reduction in time steps found at the cliff over the baseline. Table 1 shows that the action-pruning exploration approach which only selects actions from the feasible set of actions, achieves a $45.2\%$ reduction in time steps spent on the cliff compared to the baseline. The LOVR agent that explores within an $\eta$-ball around the landmark nearest to the goal state results in a $94.03\%$ reduction in time steps spent on the cliff, while the agent that combines the LOVR implementation with action-pruning resulted in a $98.95\%$ reduction. To the best of our knowledge, these results are the first to report a transfer mechanism that confers notions of safety by solely using the value functions of previously solved tasks. Moreover, this safety mechanism makes no particular assumptions on the environment itself, only that a landmark covering has previously been learned using the action-penalty reward structure, which can be implemented independantly of any endogenous reward function provided by an environment.

\begin{figure}[h!]
\centering
\caption{Cliff-Walk Domain. Red: Cliff states; Green: Start state; Blue: Goal state; Yellow: Landmark states. }
\includegraphics[width = 6cm, height =3.5cm]{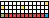}
\end{figure}

\begin{table}[h!]
   \centering
   \caption{Mean percent reduction in time-steps on the cliff relative to baseline.}
   \label{tab:table1}
   \begin{tabular}{ccc}
      \hline
      Action-Pruning & LOVR & Action-Pruning $+$ LOVR\\
	\hline
	$45.19$ & 94.03 & 98.95\\
      \hline
    \end{tabular}   
\end{table}

\subsection{MNIST-world experiments}

\paragraph{}The next set of experiments evaluate the other proposed transfer mechanisms in the multi-task RL setting. For these experiments we use a MNIST-world domain, where at each time step the agent receives three concatenated MNIST digit images, two representing the index of the x-coordinate and the third representing the index of the y-coordinate, of a $100\times 10$ grid world. At each time step the appropriate MNIST images are sampled from the training set of digits from 0-9 in order to build the representation of the current state in MNIST-world. The agent can move in each cardinal direction, where each action moves it in the appropriate cardinal direction stochastically with probability $p=0.85$, and in the other directions with equal probability. There is also a sticky absorbing state at state $(55,5)$ where the probability that an action taken is successfull is $0.05$, and with probability $0.95$ the agent remains stuck in that state. For all experiments a landmark is constructed using the $d_{rt}$ metric, with $\eta = 10, 20$. The landmark covering was built by initiating $\mathbb{L} = \empty$, then stochastically adding a single state to $\mathbb{L}$ until the state space was covered. Ultimately 27 landmark states made up the covering. The landmark centric value functions were represented using Deep Q-Networks (DQN) for each $\ell \in \mathbb{L}$, then using the landmark covering to perform the appropriate transfer mechanism. Transfer mechanism experiments involves a sequence of 20 novel goal tasks, each run for 1000 episodes, each episode lasting a maximum of 500 time steps. Each experiment was repeated a total of 5 times and we report the mean over the five experiments, with figures plotting $+/-$ 1 standard deviation. For the DQN architectures: For computational efficiency, first an autoencoder was used trained to reconstruct the MNIST dataset, where the bottleneck layer was of dimension 20. These 20-dimensional representations of the MNIST images were used as inputs to the DQN, hence each state comprises of 3 concatenated 20-dimensional vectors. The DQN itself was a simple multi-layer perceptron with two hidden layers of 200 units each, with a linear ouput layer of size $\vert \mathcal{A} ]vert = 4$. An experience replay buffer of size 25000 was used, sgd was used to train the network with mini-bathces of size 512, with ADAM optimizer with initial learning rate of $10^{-4}$.

\begin{figure}[h!]
\centering
\caption{Mean Lifelong Regret Learning Curves. Top: $\eta = 10$, Bottom: $\eta=20$. Yellow: Baseline DQN; Blue: $r^{\mathbb{L}}$; Green: LOVR DQN; Red: $\pi^{\mathbb{L}}_{\mathcal{Z}}$; Black: Bandit}
\includegraphics[width = 12cm, height =6cm]{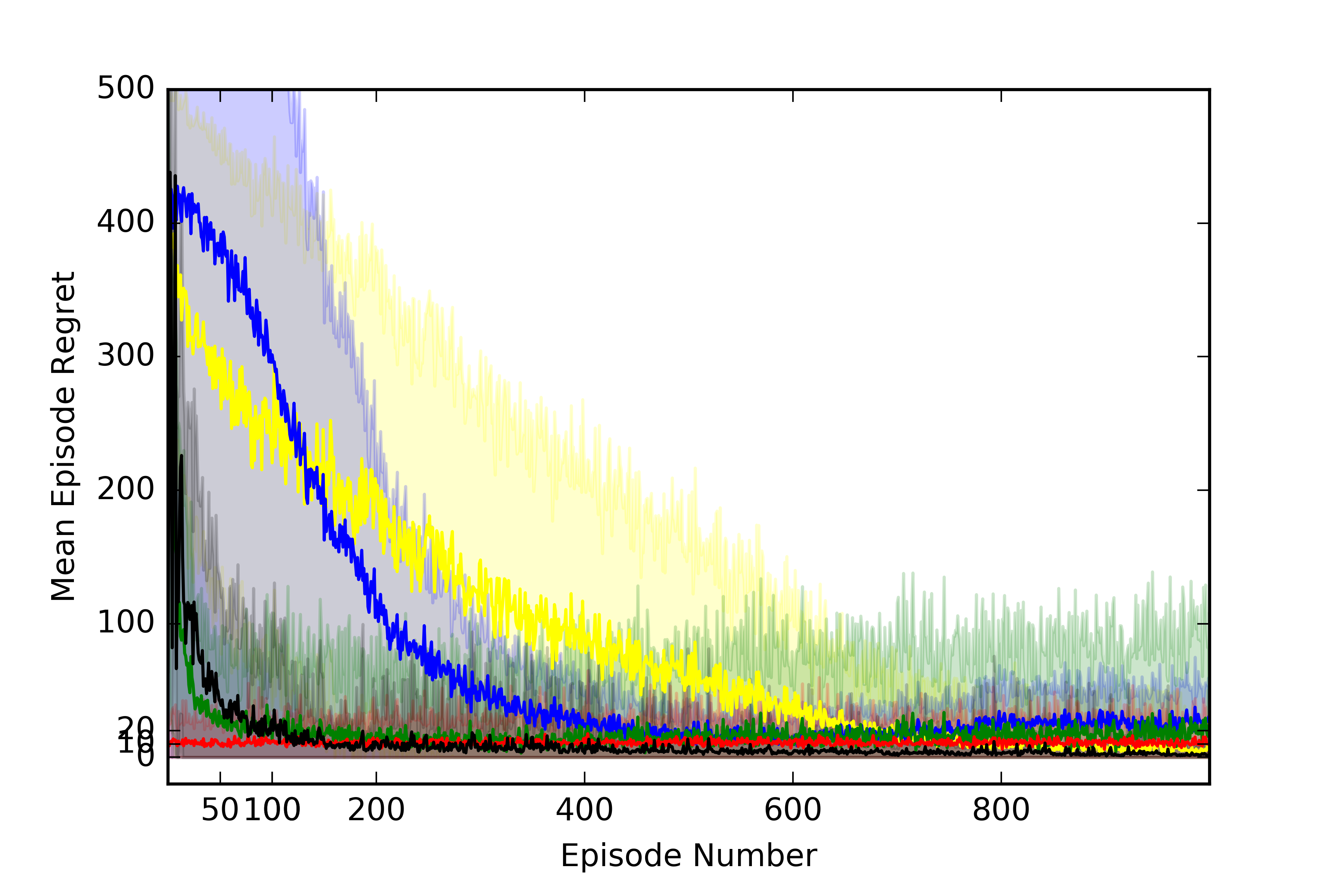}
\includegraphics[width = 12cm, height =6cm]{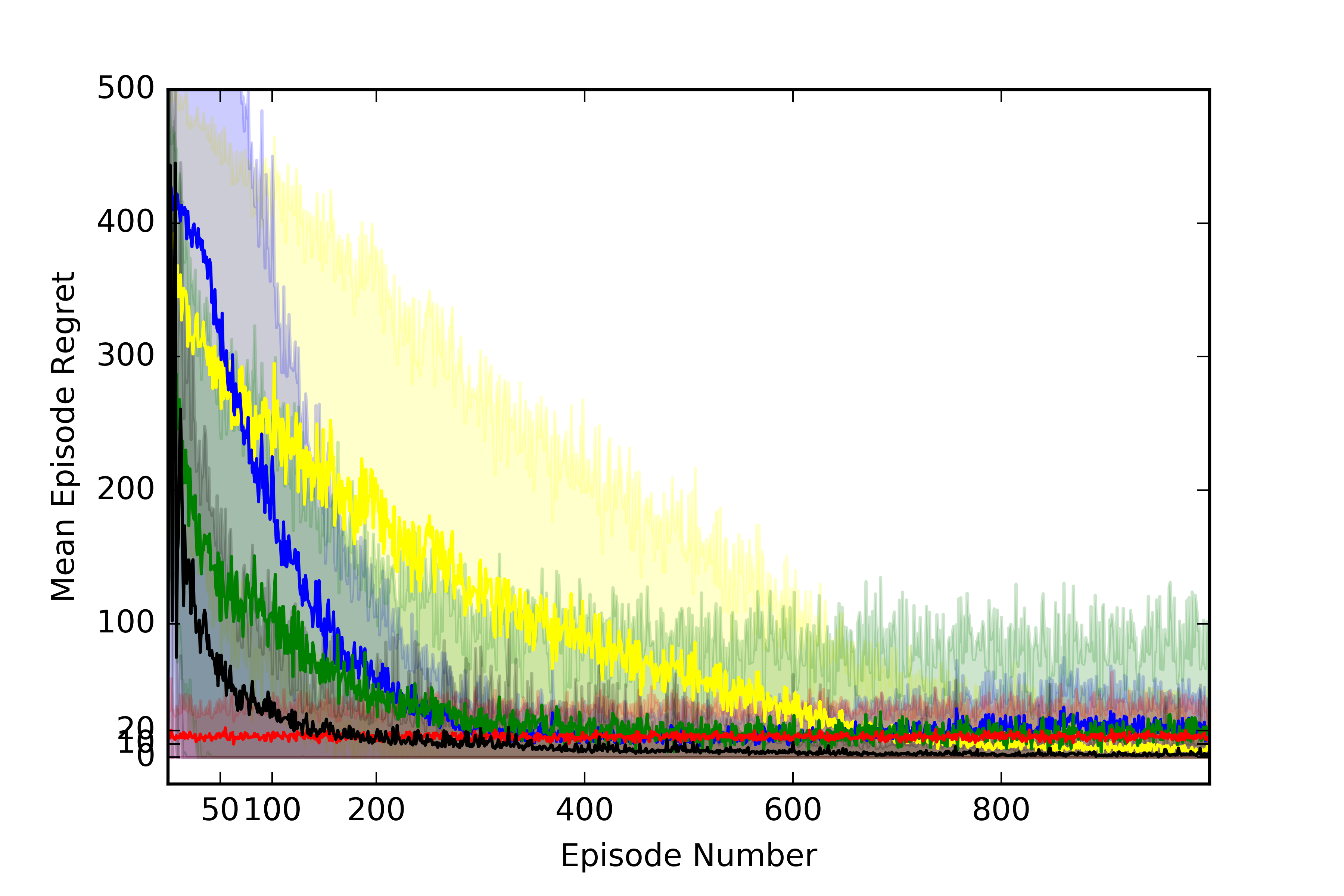}
\end{figure}

\paragraph{}Figure 2 shows the mean epsiode regret of each agent implementation, where regret is computed with respect to the optimal (oracle) value functions for each goal, given the randomly sampled initial state. As seen in Figure 2, for both $\eta = 10$ and $\eta = 20$, quite similar trends are observed. First, we note that the baseline DQN agent is slower to convergence than the other agent implementations, however asymptotically outperforms all other implementations except the bandit agent. As well, the baseline DQN agent experiences the highest variance implementation. LOVR agent (green) performs quite well, converging to a good, but not optimal policy within roughly 100 episodes. The lack of optimality is due to the inductive bias encoded in the LOVR framework which forces the agent to drive towards the nearest landmark first, before taking actions towards the current goal state. This asymptotic inefficiency is discussed in [12]. Empirically, we find that the LOVR agent solves for roughly a $2\eta$-optimal policy quite quickly, which is in line with the theoretical results previously presented [12].

\paragraph{}The zombie policy $\pi^{\mathbb{L}}_{\mathcal{Z}}$ performs exceptionally well. It is worth repeating that \textit{no learning updates whatsoever occur during these experiments}, and hence is a purely zero-shot transfer approach. The zombie policy agent uses the $V^{\mathbb{L}}$ state representation and deterministically takes actions as described above. The zombie agent achieves near oracle levels of regret in this zero-shot transfer setting, achieving an average per episode regret of 15.4 time steps as compared to an oracle policy. It can be seen that it takes the baseline DQN agent roughly 700 episodes of learning to finally overtake the $V^{\mathbb{L}}$ zombie agent in terms of mean per-episode regret. An interesting second set of experiments would have been to continue to run both against to see how many episodes it would take in order for the baseline DQN agent to overtake the zombie agent in terms of \textit{cumulative} life long regret, and not just per episode average regret. However, assuming that after episode 1000 the baseline DQN agent achieved zero regret per episode and the zombie agent continued with an average per-episode regret of 15.4, it would take over 5385 episodes for the baseline DQN agent to overtake the zombie agent with respect to cumulative life long regret under these experimental conditions.

\paragraph{}A closer look at the average per-episode regret of the $V^{\mathbb{L}}$ zombie agent is visualized in Figure 3. Upon closer inspection, we found that of the 20 task, 10 achieve better mean per-episode regret than a fully converged baseline DQN agent with 1000 episodes of learning, despite the $V^{\mathbb{L}}$ zombie agent being a purely zero-shot transfer implementation, and experiencing as little as a mean episode regret of 1.5 time steps for some tasks. For 3 of the 25 tasks, the $V^{\mathbb{L}}$ zombie agent achieved high per episode regret in the 20-28 time step range. Further analysis determined this to be attributed to the fact that for these three tasks, there were some initial states that the agent could not solve the goal from, and hence always achieved a maximum level of regret for those episodes. Despite being unable to learn from these mistakes, the $V^{\mathbb{L}}$ zombie agent still has an extremely strong performance, on average, across all 25 tasks. These results demonstrate that the $V^{\mathbb{L}}$ zombie agent may be an attractive finite time zero-shot transfer agent, when the number of episodes the agent is assigned a task is small, thus limiting the ability to learn an optimal policy quickly. Finally we see that the $r^{\mathbb{L}}$ reward transfer agent learned slightly faster than the baseline DQN agent, however it converged to a policy of poorer quality than the other agents.

\begin{figure}[h!]
\centering
\caption{Breakdown of mean per-episode regret for each of the 25 tasks for the  $V^{\mathbb{L}}$ zombie agent. Violin plot (Left) and CDF (right) to represent the density of mean per-episode regret over the 25 zero-shot transfer tasks.}
\includegraphics[width = 6.9cm, height =5cm]{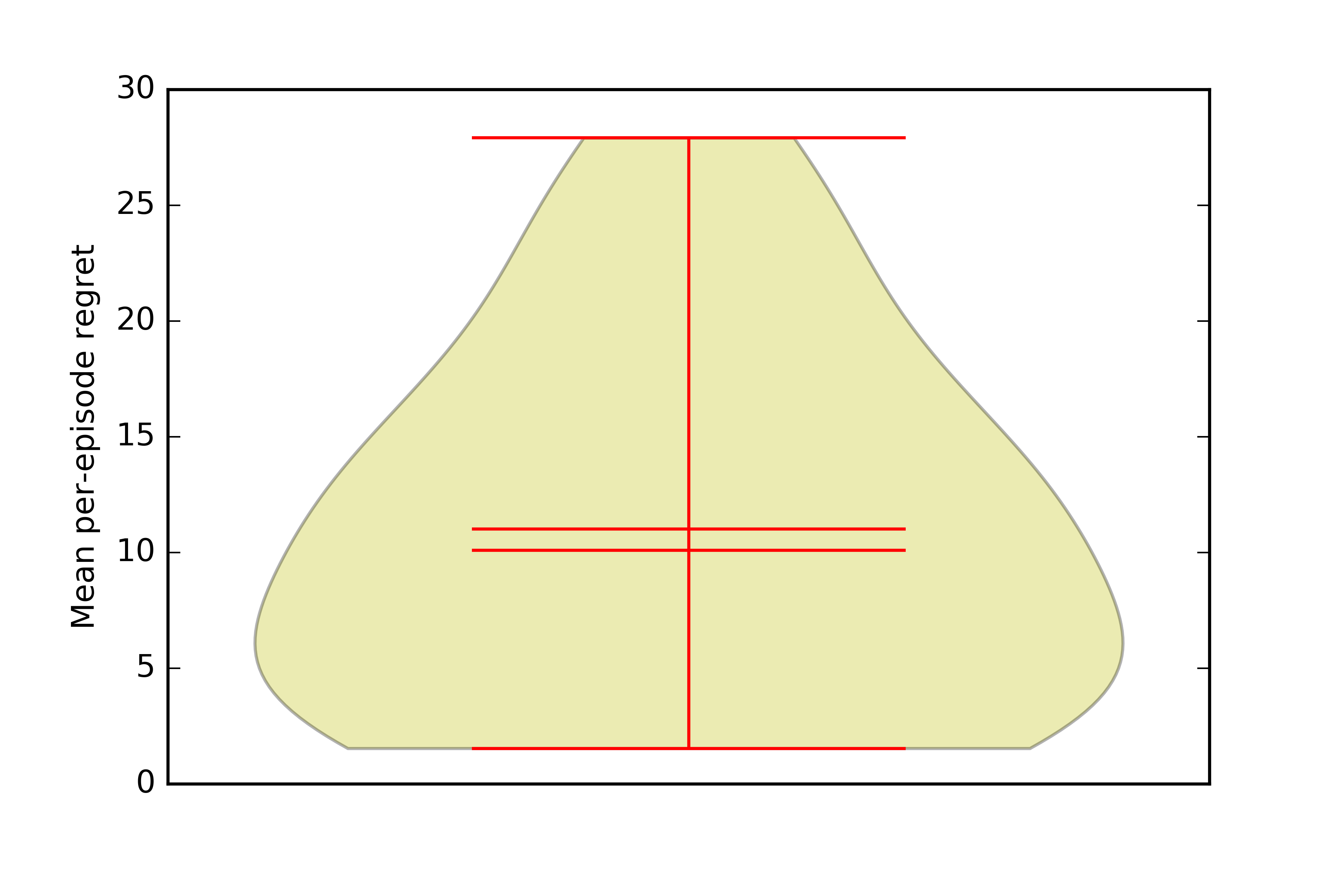}
\includegraphics[width = 6.9cm, height =5cm]{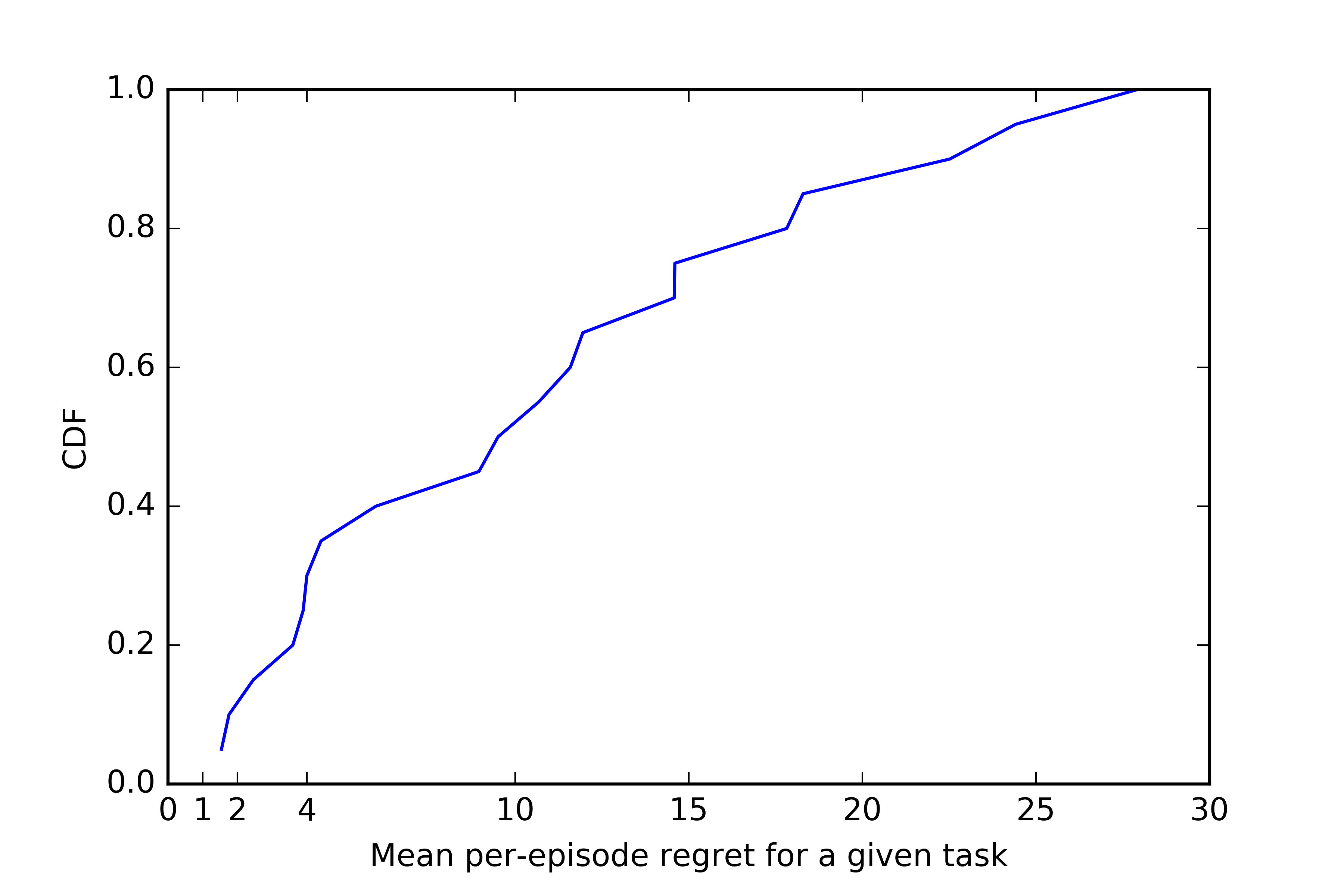}
\end{figure}

\paragraph{}Each of the transfer mechanisms tested, including the no-transfer baseline DQN agent, each have their own strengths and perform well under different conditions and at different points during throughout learning. For example, the LOVR agent begins performing quite well quickly, but is outperformed by the baseline DQN agent asymptotically. With this in mind, we implemented a bandit agent, which uses a simple multi-armed bandit controller, which selects which agent implementation to use at the start of each episode. The experience of each agent selected is used to train off-policy all the agents considered. In this way, we expect the bandit controller agent to reap the benefits of each transfer mechanism, and learn when to use each. To test this hypothesis we redo the previous experiments with a simple bandit controller as just described. The bandit controller follows the UCB algorithm with hyperparameter $C = 200$, where the reward of an arm-pull is the negative number of steps taken by the arm. Here, each transfer mechanism, including the baseline DQN agent, is an arm. The results can be seen in Figure 2 (black). In both $\eta = 10, 20$ the bandit agent solves for the optimal policy of each task quite fast, quickly overtaking the $V^{\mathbb{L}}$ zombie agent and outperforming the baseline DQN asymptotic policy in as few as 300 episodes (versus 1000 for the baseline). To investigate when the bandit controller selects each of the arms, we plot the relative proportion each arm is pulled during the course of learning (Figure 4) for experiments with $\eta =10$. The plot is noisy in for the first four episodes, since the nature of the bandit algorithm is to deterministically select each of the arms in sequence until they have all been selected once, before begging the UCB algorithm. Without considering this initial arm selection phase, it can be seen that on average across the 20 tasks, the bandit controller tends to select the LOVR agent (red) and the $V^{\mathbb{L}}$ zombie agent (green) early on in learning during approximately the first 150 episodes. The $r^{\mathbb{L}}$ agent (blue) is consistently selected the least, even less frequently than the baseline DQN agent (black). These results are consistent with the regret curves found in Figure 2. Beginning at around 150 episodes, the baseline DQN consistently becomes consistently selected more frequently by the bandit controller, along with a gradual decrease in the selection of the $V^{\mathbb{L}}$ zombie agent, whereas the LOVR agent selection frequency remains steady until roughly 250 episodes, afterwhich it gradually decreases. We see that during episodes around 600-800, the bandit controller is essentially selecting the baseline DQN agent and the $V^{\mathbb{L}}$ zombie agents equally as often, and that this corresponds to when the baseline DQN tends to overtake all other transfer agents mean per-episode regret as seen in Figure 2. Finally, the last (approximately) 200 episodes see a slight increase  in selection of the baseline DQN agent over the $V^{\mathbb{L}}$ zombie agent by the bandit controller. The bandit controller arm pull proportions are consistent with the previous experiments and as hypothesized. The bandit controller leveraged the ``fast to poor well" transfer agent implementations by favoring them early in learning, but gradually increased the selection of the asymptotically optimal baseline DQN agent. These results support the notion that each landmark convering transfer mechanism presented confers benefits at different stages of the learning process, and each with their own inductive biases favor learning at different stages.

\begin{figure}[h!]
\centering
\caption{Bandit Controller Arm Selection. Red: LOVR arm; Black: Baseline DQN arm; Green: $V^{\mathbb{L}}$ zombie arm; Blue: $r^{\mathbb{L}}$ arm.}
\includegraphics[width = 12cm, height =6cm]{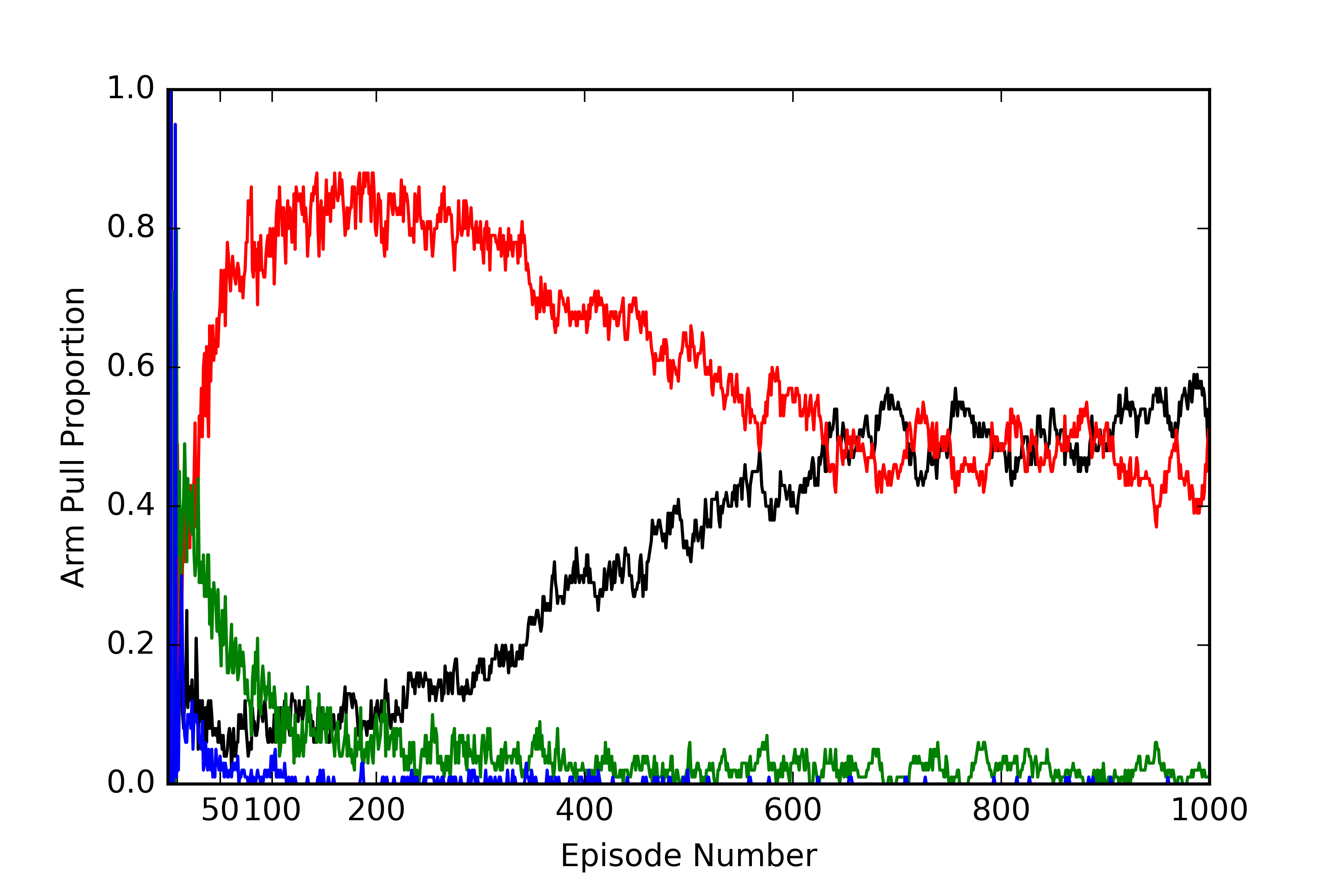}
\end{figure}

\vspace{-.5em}
\section{Discussion}

\paragraph{}We presented two novel metrics on the state space of a Markov Decision Processes, and studied various benefits, both theoretical and empirical, a topological covering under such metrics can confer to an agent in the multi-task goal based RL setting. Such a covering was introduced as a set of landmarks, $\mathbb{L}$, and was an immediate extension to the Landmark Options Via Reflection (LOVR) framework. Developing such a covering can be quite general in nature, which can be learned in a self-supervised manner, in an online fashion (adding landmark states to $\mathbb{L}$ until the state space is covered) or selected by a researcher. We show how the landmark value functions from such a topological covering can be used to confer benefits to transfer learning under the multi-task goal based RL setting. First, we explored the notion of safety and showed that the landmark value functions can be used to prune actions that may be dangerous, or infeasible with respect to the optimal policy for the current task. Empirical results on the cliff-walker domain supported the potential safety benefits landmark value functions can confer a multi-task goal based RL agent. Second, an extension to the LOVR framework that uses landmark value functions as landmark options was introduced, which had strong empirical performance. Third, motivated from the phenomenology of Heidegger, we introduced a novel state representation which uses the landmark value functions themselves as a state representation, and coupled this representation to a greedy heuristic policy called the \textit{zombie} agent. This agent was a purely zero-shot transfer mechanism and performed incredibly well on 25 novel tasks unseen by the agent. Finally, a transfer mechanism using a learned reward function based on the $V^{\mathbb{L}}$ state representation was introduced.

\subsection{Pre-print, work in progress}
\paragraph{}Landmark coverings and landmark value functions clearly carry strong empirical and theoretical [12] benefits to transfer learning in the multi-task goal based RL setting. This document is acting as a placeholder, and represents work in progress. The theoretical aspects of the $V^{\mathbb{L}}$ representation and the zombie agent require much further exploration. Initial experiments on randomly generated MDPs that are highly connected show that the zombie agent performs quite poorly, and is not meant to be touted as a general mechanism that \textit{should} or \textit{will} work for any given environment. However, some initial work and conjectures are that for state spaces with low connectivity, such as grid world environments or video games, the zombie agent should confer benefits. We also leave a more thorough discussion of relevant studies and recent research for a later draft of this paper, as well as a comparison to other baselines such as Successor Representations (SRs) which we believe also has similarities to the $V^{\mathbb{L}}$ representation.



\section{References}

[1] D. Silver, Q. Yang, and L. Li. Lifelong machine learning systems: beyond learning algorithms. In
AAAI Spring Symposium: Lifelong Machine Learning, pages 49–55, 2013.

[2] J. Baxter. A model of inductive bias learning. Journal of Artificial Intelligence Research, 12:149–198,
2000.

[3] A.G. Barto and S. Mahedevan. Recent advances in hierarchical reinforcement learning. Discrete
Event Dynamic Systems, 13:341–379, 2003.

[4] E. Brunskill and L. Li. Sample complexity of multi-task reinforcement learning. In Conference on
Uncertainty in Artificial Intelligence (UAI), 2013.

[5] K. Frans, J. Ho, P. Abbeel, and J. Schulman. Meta learning shared hierarchies. Technical report,
2017. https://arxiv.org/pdf/1710.09767[cs.LG].

[6] R. Laroche, M. Fatemi, J. Romoff, and H. van Seijen. Multi-advisor reinforcement learning. Technical
report, 2017. https://arxiv.org/pdf/1704.00756[cs.LG].

[7] H. van Seijen, M. Fatemi, J. Romoff, and R. Laroche. Separation of concerns in reinforcement
learning. Technical report, 2017. https://arxiv.org/pdf/1612.05159[cs.LG].

[8] Y. Bengio, A. Courville, and P. Vincent. Representation learning: a review and new perspectives.
Technical report, 2014. https://arxiv.org/pdf/1206.5538[cs.LG].

[9] X. Zhu. Semi-supervised learning literature survey. Technical report, 2005. Technical Report 1530.

[10] M. Fraser. Multi-step learning and underlying structure in statistical models. In NIPS, pages
4815–4823, 2016.

[11] J. Godwin, P. Stenetorp, and S. Riedel. Deep semi-supervised learning with linguistically motivated se-
quence labelling task hierarchies. Technical report, 2016. https://arxiv.org/pdf/1612.09113[cs.CL].

[12] N. Denis and M. Fraser. Options in multi-task reinforcement learning. In 32nd Canadian Conference
on Artificial Intelligence, pages 225–237, 2019.

[13] R.S. Sutton and A.G. Barto. Reinforcement Learning: An Introduction. MIT Press, 2016.

[14] S. Koenig and R.G. Simmons. Complexity analysis of real-time reinforcement learning. AAAI, pages
99–105, 1993.

[15] S.R. Sutton, D. Precup, and S. Singh. Beteween mdps and semi-mdps: a framework for temporal
abstraction in reinforcement learning. Artificial Intelligence, 112:181–211, 1999.

[16] T.A. Mann, S. Mannor, and D. Precup. Approximate value iteration with temporally extended actions.
Journal of Artificial Intelligence Research, 53:375–438, 2015.

\end{document}